\documentclass[11pt]{article}

\usepackage[final]{acl}
\usepackage{times}
\usepackage{latexsym}
\usepackage[T1]{fontenc}
\usepackage[utf8]{inputenc}
\usepackage{microtype}
\usepackage{inconsolata}
\usepackage{amsmath}
\usepackage{amssymb}
\usepackage{booktabs}
\usepackage{array}
\usepackage{enumitem}
\usepackage{graphicx}
\usepackage{multirow}
\usepackage{tikz}
\usepackage[most]{tcolorbox}
\usepackage{colortbl}
\tcbuselibrary{breakable, skins}
\usepackage{cleveref}

\usetikzlibrary{positioning, arrows.meta, shapes.geometric, fit, calc}

\newcolumntype{L}[1]{>{\raggedright\arraybackslash}p{#1}}
\newcolumntype{C}[1]{>{\centering\arraybackslash}p{#1}}

\colorlet{exampleborder}{black!35}
\colorlet{examplefill}{gray!5}

\title{Diagnosing Fine-Grained Inconsistency Classification in Financial Disclosure Text}

\author{
  Aman Kumar, Lasitha Vidyaratne, Dipanjan D Ghosh, Arnab Chakrabarti, Ahmed K Farahat \\
  Hitachi America, Ltd. 
}

\begin{document}
\maketitle

\begin{abstract}
Financial disclosures may contain numerical, temporal, referential, factual, and policy inconsistencies that require different evidence and reasoning to diagnose. We study \emph{fine-grained inconsistency classification}: given a passage known to contain a conflict, the goal is to identify its type among 11 categories. Using a fixed snapshot of the synthetic SBID-FD benchmark, we compare frozen and fine-tuned encoders, evidence-augmented classifiers, prompted large language models, and LoRA-adapted generative models under a shared evaluation protocol. Task-specific adaptation yields large improvements over frozen representations, and a fine-tuned 300M encoder performs competitively with substantially larger prompted and adapted models. We further study whether localizing the conflicting claims improves classification through matched predicted-span, reference-span, and distractor-span conditions. The results show that automatically extracted evidence provides additional signal but recovers only part of the benefit obtained from reference spans. Per-class and confusion analyses further reveal that some inconsistency types are especially sensitive to localization quality, whereas others remain difficult even when the relevant evidence is supplied. These findings identify evidence localization and fine-grained type discrimination as distinct challenges and show that compact supervised encoders are strong baselines for this task.
\end{abstract}

%%
%% 1. Introduction
%%
\section{Introduction}

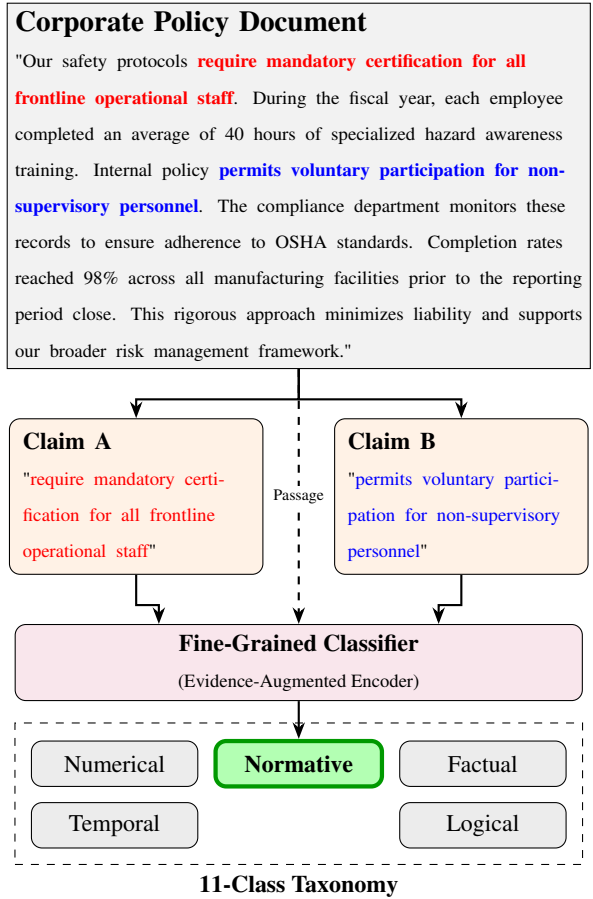
\begin{figure}[htbp]
    \centering
    \begin{tikzpicture}[
        box/.style={draw, rectangle, rounded corners, align=center, minimum height=0.6cm, fill=purple!10},
        doc/.style={draw, rectangle, align=left, text width=7.5cm, fill=gray!10, inner sep=3pt},
        span/.style={draw, rectangle, rounded corners, align=left, text width=3cm, fill=orange!10, inner sep=5pt},
        arrow/.style={-{Stealth[length=2.2mm]}, thick},
        classbox/.style={draw, rectangle, rounded corners, align=center, minimum width=2.2cm, minimum height=0.6cm, fill=gray!15, font=\footnotesize}
    ]

    % 1. Input Text Document (Full unshortened text)
    \node[doc] (input) {
        \textbf{Corporate Policy Document}\\
        % \vspace{1pt}
        \scriptsize \linespread{0.1}\selectfont
        "Our safety protocols \textbf{\textcolor{red}{require mandatory certification for all frontline operational staff}}. During the fiscal year, each employee completed an average of 40 hours of specialized hazard awareness training. Internal policy \textbf{\textcolor{blue}{permits voluntary participation for non-supervisory personnel}}. The compliance department monitors these records to ensure adherence to OSHA standards. Completion rates reached 98\% across all manufacturing facilities prior to the reporting period close. This rigorous approach minimizes liability and supports our broader risk management framework."
    };

    % 2. Extracted Evidence Spans (Side-by-side, spaced using xshift for perfect centering)
    \node[span, below=0.65cm of input.south, xshift=-2.15cm] (claimA) {
        \textbf{\footnotesize Claim A}\\
        \scriptsize  "\textcolor{red}{require mandatory certification for all frontline operational staff}"
    };

    \node[span, below=0.65cm of input.south, xshift=2.15cm] (claimB) {
        \textbf{\footnotesize Claim B}\\
        \scriptsize "\textcolor{blue}{permits voluntary participation for non-supervisory personnel}"
    };

    % 3. Classifier Model (Positioned below Claim A but shifted to the center)
    \node[box, below=0.65cm of claimA.south, xshift=2.15cm, minimum width=7.5cm] (classifier) {
        \textbf{\footnotesize Fine-Grained Classifier}\\
        \scriptsize (Evidence-Augmented Encoder)
    };

    % 4. Output Categories (Grid)
    \node[classbox, fill=green!30, draw=green!60!black, ultra thick, below=0.5cm of classifier] (norm) {\textbf{Normative}};
    \node[classbox, left=0.2cm of norm] (num) {Numerical};
    \node[classbox, right=0.2cm of norm] (fact) {Factual};
    
    \node[classbox, below=0.2cm of num] (temp) {Temporal};
    \node[classbox, below=0.2cm of fact] (logical) {Logical};

    % Grouping box for Taxonomy
    \node[fit=(temp)(num)(norm)(fact)(logical), draw, dashed, inner sep=6pt, label={[font=\footnotesize]below:\textbf{11-Class Taxonomy}}] (taxonomy) {};

    % Arrows (Input to Claims)
    \draw[arrow] (input.south) -- ++(0,-0.4) -| (claimA.north);
    \draw[arrow] (input.south) -- ++(0,-0.4) -| (claimB.north);
    
    % Arrow (Input directly to Classifier - Passage Context)
    \draw[arrow, dashed] (input.south) -- (classifier.north) node[midway, fill=white, inner sep=1.5pt] {\tiny Passage};
    
    % Arrows (Claims to Classifier)
    \draw[arrow] (claimA.south) -- ++(0,-0.4) -| ([xshift=-1.85cm]classifier.north);
    \draw[arrow] (claimB.south) -- ++(0,-0.4) -| ([xshift=1.85cm]classifier.north);

    % Arrow (Classifier to Final Prediction)
    \draw[arrow] (classifier.south) -- (norm.north);

    \end{tikzpicture}
    % \vspace{0cm}
    \caption{Fine-grained inconsistency classification utilizing paired evidence spans.}
  \label{fig:teaser}

\end{figure}

Financial disclosures combine quantitative statements, temporal claims, entity
references, policy commitments, legal qualifications, risk descriptions, and
forward-looking projections. These statements are often produced and revised
across different teams, data sources, and reporting cycles, creating
opportunities for conflicts that differ substantially in form and in the
reasoning required to diagnose them. A numerical inconsistency may require
checking arithmetic or tabular aggregation; a temporal inconsistency may require
comparing reporting periods or event sequences; a referential inconsistency may
require tracking entities across clauses; and a normative inconsistency may
require determining whether stated obligations or policies are mutually
compatible. Inconsistency in this setting is not a single homogeneous
phenomenon: different types depend on different evidence, reasoning patterns, and
downstream remediation steps. Figure~\ref{fig:teaser} depicts a passage with a normative conflict, its
conflicting claims, and the typing decision.

Financial text presents domain-specific challenges that generic models struggle
to handle. Its lexical and semantic behavior diverges from everyday usage: general-purpose sentiment
dictionaries can misclassify common financial terms in 10-K filings
\citep{loughran2011liability}, while domain-specific models have
shown benefits in both financial and other technical domains
\citep{liu2020finbert,kumar2026building}. Financial
question-answering benchmarks further show that understanding financial reports
often requires reasoning over heterogeneous textual and tabular evidence,
including numerical operations and multi-step derivations
\citep{chen2021finqa,zhu2021tatqa}. These studies motivate financial disclosures
as a domain in which generic language understanding is insufficient, but they
target sentiment analysis, representation learning, or question answering rather
than fine-grained inconsistency typing.

% \documentclass{article}
% \usepackage{tikz}
% \usetikzlibrary{positioning, shapes.geometric, arrows.meta, fit, backgrounds, calc}
% \usepackage{geometry}
% \usepackage{xcolor}
% \geometry{a4paper, margin=1in}

% \begin{document}

% \end{document}

Prior work has studied inconsistency and contradiction primarily through natural
language inference (NLI), fact verification, and evidence-grounded prediction.
Large-scale NLI benchmarks such as SNLI, MultiNLI, and ANLI frame contradiction
as a relation between a premise and a hypothesis, assigning one of entailment,
contradiction, or neutral \citep{bowman2015large,williams2018broad,williams2020anli}.
Fact-verification benchmarks such as FEVER instead ask whether a claim is
supported, refuted, or unverifiable with respect to evidence
\citep{thorne2018fever}, while rationale-based evaluation work such as ERASER
studies whether model predictions can be grounded in supporting text spans
\citep{deyoung2020eraser}. More recent resources extend inference beyond sentence
pairs: ContractNLI applies document-level inference and evidence identification
to contracts \citep{koreeda2021contractnli}, and ContraDoc studies
self-contradictions within long documents \citep{li2024contradoc}. Most closely related, FIND evaluates whether models can locate expert-inserted
inconsistencies in long financial documents, requiring both the supporting
spans and an explanation of the conflict \citep{lovering2026finding}. That work
targets detection and free-form explanation, using inconsistency type as an
analysis variable, whereas we treat typing as the prediction task itself. These lines of
work establish contradiction detection, claim verification, and evidence
grounding as central problems, but they do not directly address the complementary
question studied here: once a passage is known to contain an inconsistency, can a
model identify which specific \emph{type} it contains?

We refer to this task as \textit{fine-grained inconsistency classification}.
The task is useful diagnostically because it reveals which inconsistency types
current models handle reliably and which remain difficult. It may also support
future review workflows by routing numerical, temporal, referential, normative,
and other conflicts toward different downstream checks. The present study does
not evaluate such a deployed workflow; instead, it uses a controlled synthetic
testbed to analyze classification behavior and the role of localized evidence.

This paper makes four contributions.

\begin{enumerate}[leftmargin=*,itemsep=2pt]

\item \textbf{Cross-model evaluation.}
We provide a systematic comparison of frozen encoders, fully adapted encoders, evidence-augmented classifiers, prompted LLMs, and LoRA-adapted models on fine-grained inconsistency typing in financial text, under one 11-class protocol.

\item \textbf{Compact supervised adaptation.}
We show that a fine-tuned 300M encoder is competitive with substantially larger
prompted and LoRA-adapted systems while requiring markedly less compute.

\item \textbf{Localization diagnostic.}
Matched predicted-span, gold-span, and distractor-span conditions
help distinguish localization-related errors from errors that remain after
the relevant evidence is supplied.

\item \textbf{Class-level error analysis.}
Per-class and confusion analyses identify categories limited primarily by
localization and those that remain difficult under gold evidence.

\end{enumerate}

%%
%% 2. Task and Data
%%
\section{Task and Data}

\begin{table}[!htbp]
\centering
\caption{Definitions of the 11 inconsistency categories.}
\label{tab:taxonomy}
\small
\begin{tabular}{p{1.6cm} L{0.68\columnwidth}}
\toprule
\textbf{Type} & \textbf{Definition} \\
\midrule
Numerical &
Incompatible quantities, totals, percentages, or calculations. \\

Unit \& Measurement &
Conflicting units, scales, currencies, or measurement bases. \\

Temporal &
Incompatible dates, periods, durations, or event orderings. \\

Factual &
Contradictory claims about an entity, event, or state of affairs. \\

Logical &
Claims whose implications cannot jointly hold. \\

Theoretical &
A claim conflicts with an applicable conceptual or domain principle. \\

Referential &
References or identifiers resolve to incompatible entities. \\

Terminological &
A term is used with incompatible meanings or definitions. \\

Specificity \& Scope &
Claims differ incompatibly in population, condition, or scope. \\

Pragmatic &
A stated action or implication conflicts with the surrounding intent. \\

Normative \& Policy &
Obligations, permissions, policies, or rules are incompatible. \\
\bottomrule
\end{tabular}
\end{table}

\paragraph{Task definition.}
Given an inconsistent input text $T_j$, the task is to predict an inconsistency
label $\hat{c}_j$ from 11 predefined categories. The prediction is correct if
$\hat{c}_j = c_j$, the ground-truth label.

The task conditions on an inconsistency already being detected and therefore
evaluates \emph{typing}, not upstream detection. False positives, false
negatives, and end-to-end detection-to-typing error propagation are outside the
scope of this study.

\paragraph{Data.}
We use a fixed early snapshot of SBID-FD, a synthetic financial-disclosure benchmark introduced in a companion paper currently under review \cite{wong2027sbid}. 
The snapshot contains 5,940 instances derived from 970 distinct base passages and predates the benchmark's expanded public release; all experiments in this paper were conducted on this snapshot, which we include with the supplementary material for exact reproducibility. Each instance contains the inconsistent passage, its inconsistency label from the 11-category taxonomy in \cref{tab:taxonomy}, an automatically generated reference explanation, and two verbatim reference spans identifying the anchoring and conflicting claims.

The data were produced through a multi-stage LLM pipeline. A pool of language models, drawn from several open-weight families, first generated internally consistent SEC-style disclosure passages across varied business contexts, with each base passage screened by an LM judge for internal coherence. A separate generation step then introduced one inconsistency of a specified target type, which serves as the gold label, while preserving the surrounding context, discourse structure, and most of the original wording. Candidate variants passed a lexical pre-filter and multiple LM-judge quality gates that reject shallow, label-misaligned, or implausible edits, after which an automatic evidence-generation step produced the reference explanation and paired evidence spans. The evidence-generation stage also serves as an additional automated validation gate by requiring the generated explanation to be grounded in verbatim passage spans. No human annotation was used at any stage. The generator pool includes Qwen3.5 models that we also evaluate as prompted classifiers, and GPT-5-mini, which shares a model family with the evaluated GPT-5.4; we return to this overlap in the Limitations. The taxonomy and generation pipeline are described in the companion benchmark paper and are used here as a fixed evaluation testbed.

The primary experiments use a stratified 70--30 train-test split containing
4,158 training and 1,782 test examples. We additionally evaluate a stricter
base-passage-disjoint split in which all variants derived from the same base
passage are assigned to only one partition
(Appendix~\ref{app:base_disjoint}).%\footnote{We will release the exact split
%identifiers used in both evaluations together with the experiment code.} 
The primary test-set
distribution ranges from 109 Referential examples to 241 Logical examples; full
per-class counts are reported in \Cref{tab:class_counts}.

The inconsistent passages are short excerpts rather than full documents, with a
median length of approximately 87 words; their corresponding base passages have
a median length of approximately 78 words. The median character-level
Levenshtein distance between a base passage and its inconsistent variant is 72,
and the median longest common contiguous substring covers approximately 73\% of
the shorter passage. These statistics show that the variants generally preserve
most of the base-passage surface form while introducing a localized edit or
insertion, enabling the evidence-localization analyses studied here. A worked
example and complete length and overlap statistics appear in
Appendix~\ref{app:examples}.

%%
%% 3. Methods
%%
\section{Methods}
\label{sec:methods}
 
We compare two families of approaches. The first is discriminative embedding-based
classification: we embed the passage (and, optionally, predicted conflicting
spans) and predict the inconsistency type with a small multilayer perceptron
(MLP) head (\Cref{fig:mlp_architecture}); this family covers the passage
classifiers and the evidence-augmented classifiers below. The second is
generative: we prompt instruction-tuned LLMs to emit the label directly, and
also fine-tune two such models with LoRA. All approaches use the same task, split,
and label set.

\subsection{Passage Classifiers}
We consider both frozen and fine-tuned passage encoders, evaluated under the
same downstream classifier to keep the comparison fair.

\paragraph{Frozen setting.}
We embed the inconsistent passage with a frozen sentence encoder and train a
two-layer MLP on top (hidden sizes [512, 256], dropout 0.2, lr $=10^{-3}$). We
compare five backbones: MiniLM-L6-v2 \citep{wang2020minilm,reimers2019sbert},
Qwen3-Embedding 0.6B and 8B \citep{zhang2025qwen3embedding},
EmbeddingGemma-300M \citep{schechter2025embeddinggemma}, and DeBERTa-v3-base
\citep{he2023debertav3} (mean-pooled).

\paragraph{Fine-tuned setting.}
Among general-purpose frozen backbones, EmbeddingGemma-300M achieves the highest accuracy
(27.1\%), making it the natural candidate for full fine-tuning. For
EmbeddingGemma-300M, fine-tuning proceeds in two steps. First, we fine-tune
the full encoder end-to-end for 11-way classification using a linear head,
adapting the encoder weights to the task. Second, we discard the linear head,
freeze the adapted encoder, precompute passage embeddings, and train a new
two-layer MLP classifier on those embeddings, using the identical MLP
architecture as the frozen baseline. The reported fine-tuned numbers come from
this second step, so the only difference from the frozen baseline is the quality
of the underlying embeddings.

To test whether the frozen result is merely an artifact of a weak frozen
backbone, we also fine-tune DeBERTa-v3-base as a passage classifier following
the same two-step protocol, so the same backbone appears in both frozen and
fine-tuned form. For each encoder we report both conditions: (i) the linear
classifier from end-to-end fine-tuning, and (ii) the MLP classifier after
freezing the fine-tuned encoder. Hyperparameters are in
Appendix~\ref{app:lora_traditional}.

We additionally evaluate FinBERT \citep{araci2019finbert}, a finance-domain-pretrained
encoder, under the identical frozen, fine-tuned, and evidence-augmented protocol,
to test whether domain-specific pretraining narrows the gap to the primary
EmbeddingGemma pipeline. We report it
alongside the other backbones in \Cref{tab:encoder_results,tab:evidence_results}.

\subsection{Evidence-Augmented Classifiers}

\begin{figure}[!htbp]
\centering
\resizebox{0.9\columnwidth}{!}{%
\begin{tikzpicture}[
    node distance=0.5cm,
    box/.style={rectangle, draw, rounded corners,
                minimum width=2.6cm, minimum height=0.6cm,
                font=\normalsize, align=center},
    mlpbox/.style={rectangle, draw, rounded corners,
                text width=3.4cm, minimum height=1.1cm,
                font=\normalsize, align=center, fill=green!10},
    arrow/.style={-{Stealth}, thick}
]
\node[box, fill=orange!10] (qa) {EmbeddingGemma QA\\{\small (extracts spans at test time)}};
\node[box, fill=orange!15, below=of qa] (ev_input) {Evidence Spans\\{\small (claim\_a, claim\_b)}};
\node[box, fill=orange!25, below=of ev_input] (ev_emb) {Span Embeddings\\{\small (EmbGemma-300M)}};
\node[box, fill=orange!35, below=of ev_emb] (ev_pool) {Mean Pooling};
\draw[arrow] (qa) -- (ev_input);
\draw[arrow] (ev_input) -- (ev_emb);
\draw[arrow] (ev_emb) -- (ev_pool);
\node[draw=orange!60, dashed, rounded corners, fit=(qa)(ev_input)(ev_emb)(ev_pool), inner sep=8pt] (ev_box) {};
\node[font=\footnotesize, color=orange!80!black, above=2pt of ev_box] {Evidence branch};

\node[box, fill=blue!10, left=3.5cm of qa] (text_input) {Inconsistent Text};
\node[box, fill=blue!20, below=of text_input] (text_emb) {Text Embedding\\{\small (EmbGemma-300M)}};
\draw[arrow] (text_input) -- (text_emb);

\node[box, fill=gray!20] at (text_input |- ev_pool) (concat) {Concatenate \ {\small text $\oplus$ evidence}};
\draw[arrow] (text_emb) -- (concat);
\draw[arrow] (ev_pool) -- (concat);

\node[mlpbox, below=0.7cm of concat] (mlp) {MLP\\[2pt]{\small 1536$\to$512$\to$256$\to$11}\\{\small ReLU, Dropout 0.2}};
\node[box, fill=red!15, below=0.6cm of mlp] (output) {Predicted Label \ {\small 11 classes}};
\draw[arrow] (concat) -- (mlp);
\draw[arrow] (mlp) -- (output);
\end{tikzpicture}%
}
\caption{Evidence-augmented classifier. An EmbeddingGemma QA model predicts the
two conflicting spans, which are pooled and concatenated with the passage
embedding before the MLP head. The passage-only classifier omits the evidence
branch; the gold-span oracle replaces the predicted spans with gold spans at
test time.}
\label{fig:mlp_architecture}
\end{figure}
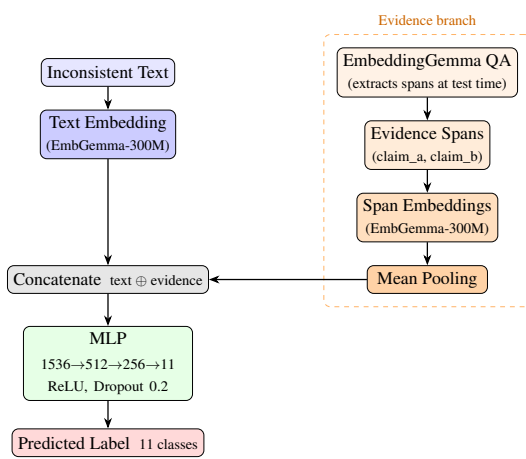

Each instance contains two reference evidence spans: \texttt{claim\_a}, which
serves as the anchoring claim, and \texttt{claim\_b}, which introduces or
completes the conflict. We use this paired-span representation as a controlled
way to study whether localized conflict evidence improves inconsistency typing.
The representation assumes that the inconsistency can be expressed through two
verbatim spans within the passage, an assumption supported by the construction
of this dataset but not guaranteed for naturally occurring financial
disclosures. At test time, we replace the reference spans with outputs from an
extractive question-answering model and pass the predicted spans to the
classifier (\Cref{fig:mlp_architecture}).

\paragraph{Span prediction.}
We frame localization as extractive QA over the passage. The model is run twice
per passage, once with a fixed role question for claim A and once for claim B,
and decodes one contiguous span per role from the start and end token logits
(the highest $\text{start}+\text{end}$ logit sum over the top-25 start and end
candidates per window, with a maximum span length of 80 tokens). Roles are
assigned by which question is posed; the model does not predict the role itself.
Each training row yields up to two QA examples, one per gold span. We compare
several encoder backbones (Appendix~\ref{app:evidence_qa}); our strongest
extractor attaches an extractive-QA head to the EmbeddingGemma-300M backbone
\citep{schechter2025embeddinggemma}, which we use in the primary predicted-span
configuration. EmbeddingGemma is released
as an embedding model that outputs a single pooled representation rather than
token-level start/end logits; we therefore discard the release pooling head and
fine-tune the underlying transformer with a standard span head under the same
QA protocol as the other backbones. This extractor attains the highest span quality among the evaluated backbones
(Token-F1 0.679, compared with 0.611 for DeBERTa-v3-base), and we select it for
the primary predicted-span experiments. The full backbone comparison, decoding
details, and span-quality results are reported in
Appendix~\ref{app:evidence_qa}; additional analyses supporting this selection
are provided in Appendix~\ref{app:gemma_extractor}.

\paragraph{Feature construction.}
Let $e(\cdot)$ denote the L2-normalized EmbeddingGemma-300M encoder. Given
spans $s_a, s_b$ (predicted at test time, gold during training), the classifier
input is
\[
\mathbf{f}
=
\left[
e(T)\,;\,
\frac{1}{2}\left(e(s_a) + e(s_b)\right)
\right]
\in \mathbb{R}^{1536},
\]
the concatenation of the passage embedding with the mean-pooled span
embeddings, fed to the same [512, 256] MLP head. We evaluate this branch in three encoder regimes: the original frozen encoder;
the fine-tuned EmbeddingGemma encoder used as a frozen feature extractor after
passage-level adaptation, which serves as our primary evidence-augmented
configuration; and a fully end-to-end variant in which the encoder receives the
passage and both spans and is trained with gradients flowing through it.

\paragraph{Matched training and the gold-span oracle.}
All evidence-augmented classifiers are trained using reference spans from the
training partition. The test conditions differ only in the source of the
evidence: the automatic condition uses QA-predicted spans, while the oracle
condition supplies reference spans. Because the encoder, classifier head, and training examples are otherwise
identical, the predicted-to-gold difference primarily reflects the effect of
test-time evidence quality. However, because the classifier is trained on
reference spans but evaluated on imperfect predicted spans, part of this gap
may also reflect sensitivity to the resulting train--test evidence mismatch.

This protocol should be interpreted as a diagnostic experiment rather than a
span-supervision-free deployment setting. It assumes access to paired
claim-level evidence during classifier and extractor training. Such supervision
may be costly or unavailable for naturally occurring financial disclosures,
and evaluating weaker or indirect supervision is outside the scope of the
present study.

\subsection{Generative Models}
We prompt three instruction-tuned LLM families to output one of the 11 labels
given the passage, label definitions, and one illustrative example per label
(deterministic decoding: temperature~$= 0$ for GPT-5.4, \texttt{do\_sample=False}
for local models; \texttt{max\_new\_tokens} set per model family: 8{,}192 for
Qwen3.5, 96 for Gemma-4, 32 for GPT-5.4; full prompt in
Appendix~\ref{app:prompt}): Qwen3.5 (4B, 9B, 27B,
35B-A3B) \citep{qwen2026qwen35}, Gemma-4 (E4B-it, 26B-A4B-it, 31B-it)
\citep{google2026gemma4}, and GPT-5.4 \citep{openai2026gpt54}. All evaluated outputs contained a single valid label after parsing, and the larger generation limits were therefore not reached.
We additionally fine-tune Qwen3.5-9B and Qwen3.5-27B with QLoRA
(quantized Low-Rank Adaptation; \citealp{hu2022lora,dettmers2023qlora}) using
supervised fine-tuning. Each example pairs an instruction-style prompt
containing the excerpt and label list with the gold inconsistency-type string as
the completion, optimizing the standard causal language-modeling objective over
the prompt-label sequence. Full hyperparameters are reported in
Appendix~\ref{app:lora}.

\paragraph{Metrics and protocol.} We report Accuracy, Macro-F1, and Weighted-F1
on the fixed test set (1{,}782 examples). For the fine-tuned encoder models we
report mean and standard deviation over three seeds (re-running encoder
fine-tuning and classifier training per seed); the data split is held fixed. The Qwen3.5-9B LoRA model is evaluated over three seeds, while the
Qwen3.5-27B LoRA result is single-seed. Frozen-encoder results are single-seed,
and prompted results are near deterministic under greedy decoding. We report multi-seed variance for the trained conditions most central to the
headline comparison; split and extractor variance are not included. Per-class F1 scores for all methods are reported in
Appendices~\ref{app:perclass_traditional}, \ref{app:perclass_finetuned},
and~\ref{app:perclass_generative}.

\section{Results}

\begin{table}[!htbp]
\centering
\caption{Encoder-based results on the 11-class inconsistency classification
task. EmbGemma = EmbeddingGemma-300M; QwenEmb = Qwen3-Embedding; FinBERT =
ProsusAI/finbert. Values with $\pm$ report mean and standard deviation over
three seeds; all other rows are single-seed.}
\label{tab:encoder_results}
\resizebox{\columnwidth}{!}{%
\begin{tabular}{l c c c}
\toprule
\textbf{Method} & \textbf{Acc.} & \textbf{Mac. F1} & \textbf{Wei. F1} \\
\midrule
\multicolumn{4}{l}{\textit{Frozen Passage Encoder + MLP}} \\
\quad MiniLM
& 0.189 & 0.157 & 0.173 \\
\quad QwenEmb-0.6B
& 0.204 & 0.163 & 0.187 \\
\quad QwenEmb-8B
& 0.233 & 0.202 & 0.227 \\
\quad DeBERTa-v3-base
& 0.234 & 0.151 & 0.178 \\
\quad EmbGemma
& 0.271 & 0.220 & 0.248 \\
\quad FinBERT
& 0.276 & 0.209 & 0.236 \\
\midrule
\multicolumn{4}{l}{\textit{Fine-Tuned Passage Encoder}} \\
\quad DeBERTa-v3-base (linear)
& 0.468 & 0.363 & 0.406 \\
\quad DeBERTa-v3-base (+ MLP)
& 0.496 & 0.434 & 0.471 \\
\quad FinBERT (linear)
& 0.593 & 0.571 & 0.584 \\
\quad FinBERT (+ MLP)
& 0.579 & 0.570 & 0.581 \\
\quad EmbGemma (linear)
& $0.611{\pm}0.010$ & $0.600{\pm}0.013$ & $0.607{\pm}0.013$ \\
\quad EmbGemma (+ MLP)
& $0.619{\pm}0.006$ & $0.614{\pm}0.005$ & $0.620{\pm}0.007$ \\
\bottomrule
\end{tabular}}
\end{table}

\subsection{Main Comparison}
\Cref{tab:encoder_results,tab:evidence_results,tab:generative_results} report
the encoder-based, evidence-augmented, and generative-model comparisons,
respectively. Uniform random guessing yields 9.1\% expected accuracy, while always predicting the majority class yields 13.5\%.

\paragraph{Fine-tuning substantially improves all three encoder backbones.}
As shown in \Cref{tab:encoder_results}, a frozen EmbeddingGemma encoder with an
MLP head reaches 27.1\% accuracy, while increasing the frozen backbone size
within the Qwen3-Embedding family yields only a modest change
(0.6B: 20.4\%; 8B: 23.3\%). Frozen DeBERTa-v3-base and FinBERT are similarly
weak, at 23.4\% and 27.6\%. Fine-tuning the same 300M EmbeddingGemma encoder
for the classification task raises accuracy to $61.9 \pm 0.6\%$ and Macro-F1
to $61.4 \pm 0.5\%$, an improvement of more than 34 absolute accuracy points.
The same pattern appears across all three backbones tested: fine-tuning
DeBERTa-v3-base raises its accuracy from 23.4\% to 49.6\% (a gain of 26.2
points), and fine-tuning FinBERT raises its accuracy from 27.6\% to 59.3\%
(a gain of 31.7 points). These controlled within-backbone comparisons show
that the frozen representations are poorly aligned with the fine-grained
label space and that task-specific supervision substantially improves their
discriminative utility, regardless of whether the backbone is general-purpose
or finance-domain-pretrained. They do not, by themselves, establish a general
relationship between adaptation and parameter scale across different model
families.

\begin{table}[!htbp]
\centering
\caption{Evidence-augmented results on the 11-class task. Same encoders,
evidence branch, and MLP head as \Cref{tab:encoder_results}; pred.\ =
predicted evidence spans. $^{\dagger}$Single-seed EmbeddingGemma-extractor
or end-to-end result.}
\label{tab:evidence_results}
\resizebox{\columnwidth}{!}{%
\begin{tabular}{l c c c}
\toprule
\textbf{Method} & \textbf{Acc.} & \textbf{Mac. F1} & \textbf{Wei. F1} \\
\midrule
\quad frozen, pred.\ (DeBERTa)
& 0.510 & 0.497 & 0.511 \\
\quad frozen, pred.\ (FinBERT QA)
& 0.512 & 0.488 & 0.506 \\
\quad frozen, pred.\ (Gemma)
& 0.532 & 0.517 & 0.531 \\
\quad frozen, gold \textit{(oracle)}
& 0.609 & 0.602 & 0.611 \\
\midrule
\quad fine-tuned, pred.\ (FinBERT QA)
& 0.590 & 0.584 & 0.596 \\
\quad fine-tuned, pred.\ (DeBERTa)
& 0.621 & 0.614 & 0.621 \\
\quad fine-tuned, pred.\ (Gemma)$^{\dagger}$
& 0.633 & 0.627 & 0.634 \\
\quad fine-tuned, gold \textit{(oracle)}
& 0.653 & 0.645 & 0.653 \\
\midrule
\quad end-to-end, pred.\ (FinBERT QA)
& 0.600 & 0.575 & 0.590 \\
\quad end-to-end, pred.\ (DeBERTa)$^{\dagger}$
& 0.623 & 0.611 & 0.620 \\
\quad end-to-end, pred.\ (Gemma)$^{\dagger}$
& 0.635 & 0.623 & 0.631 \\
\quad end-to-end, gold \textit{(oracle)}$^{\dagger}$
& 0.674 & 0.665 & 0.670 \\
\bottomrule
\end{tabular}}
\end{table}

\paragraph{FinBERT performs comparably to EmbeddingGemma, without an 
advantage.} FinBERT \citep{araci2019finbert} reaches 27.6\% frozen and 59.3\%
fine-tuned accuracy, close to EmbeddingGemma's 27.1\% and $61.9\pm0.6\%$; with
predicted spans it reaches 59.0\% vs.\ EmbeddingGemma's 63.3\%, reflecting its
weaker QA extractor (Token-F1 0.613 vs.\ 0.679). Domain-specific pretraining
doesn't yield a clear advantage here.

\paragraph{A compact supervised encoder is competitive with much larger
models.}
Comparing \Cref{tab:encoder_results,tab:generative_results}, the fine-tuned
300M encoder reaches $61.9 \pm 0.6\%$ accuracy, compared with
$61.5 \pm 0.2\%$ for Qwen3.5-9B with LoRA and 61.3\% for prompted GPT-5.4.
Increasing the LoRA-adapted model from 9B to 27B parameters raises accuracy to
62.9\% in a single run, while the prompted open-weight models remain below the
fine-tuned encoder. These results do not constitute a controlled scaling study:
the encoder, prompted LLMs, and LoRA-adapted models differ in architecture,
training objective, supervision format, prompting, and decoding. Nevertheless,
they provide a practically relevant comparison under the same dataset and label
space: the compact supervised encoder achieves accuracy comparable to the
strongest substantially larger systems tested while requiring markedly less
training time and compute. For context, fine-tuning the
EmbeddingGemma passage encoder takes approximately four minutes per seed on a
single RTX~6000 Ada GPU, compared with approximately 50 minutes for
Qwen3.5-9B LoRA and 2.5 hours for Qwen3.5-27B LoRA; full compute details are
reported in Appendix~\ref{app:lora_traditional}.

\begin{table}[!htbp]
\centering
\caption{Prompted and LoRA-adapted generative-model results on the same
11-class task. Prompted models use label definitions and one illustrative
example per label. The Qwen3.5-9B LoRA result reports mean and standard
deviation over three seeds; all other rows are single-seed.
$^{\dagger}$Single-seed LoRA result.}
\label{tab:generative_results}
\resizebox{\columnwidth}{!}{%
\begin{tabular}{l c c c}
\toprule
\textbf{Method} & \textbf{Acc.} & \textbf{Mac. F1} & \textbf{Wei. F1} \\
\midrule
\multicolumn{4}{l}{\textit{Prompted Generative}} \\
\quad Qwen3.5-4B
& 0.274 & 0.244 & 0.256 \\
\quad Qwen3.5-9B
& 0.330 & 0.275 & 0.289 \\
\quad Qwen3.5-35B-A3B
& 0.389 & 0.324 & 0.349 \\
\quad Gemma-4-31B-it
& 0.404 & 0.366 & 0.385 \\
\quad Gemma-4-E4B-it
& 0.409 & 0.397 & 0.414 \\
\quad Gemma-4-26B-A4B-it
& 0.439 & 0.411 & 0.434 \\
\quad Qwen3.5-27B
& 0.487 & 0.472 & 0.482 \\
\quad GPT-5.4
& 0.613 & 0.589 & 0.600 \\
\midrule
\multicolumn{4}{l}{\textit{LoRA-Adapted Generative}} \\
\quad Qwen3.5-9B + LoRA SFT
& $0.615{\pm}0.002$ & $0.596{\pm}0.002$ & $0.604{\pm}0.003$ \\
\quad Qwen3.5-27B + LoRA SFT$^{\dagger}$
& 0.629 & 0.610 & 0.622 \\
\bottomrule
\end{tabular}}
\end{table}

\paragraph{Predicted conflict spans can provide additional classification signal.}
As shown in \Cref{tab:evidence_results}, on the primary split, adding predicted
spans raises the fine-tuned encoder from 61.9\% to 63.3\% accuracy and from
61.4\% to 62.7\% Macro-F1. Relative to the generative results in
\Cref{tab:generative_results}, the resulting model exceeds Qwen3.5-9B with
LoRA (61.5\%) and prompted GPT-5.4 (61.3\%), and is comparable to the
single-run Qwen3.5-27B LoRA result (62.9\%). End-to-end passage-and-span
training reaches 63.5\% with predicted evidence and 67.4\% with gold evidence.
Thus, localized evidence can complement the passage representation, but the
larger gold-span gain indicates unrealized benefit from more accurate
extraction.

\paragraph{Base-passage-disjoint robustness.}
To test whether the main findings depend on shared base-passage scaffolds
across train and test, we retrain the fine-tuned EmbeddingGemma passage
classifier and matched evidence-augmented variants on a stricter split that
assigns all variants of each base passage to a single partition
(test $n{=}1{,}813$). The passage-only, predicted-span, and gold-span models
reach 63.0\%, 63.0\%, and 65.0\% accuracy, respectively, preserving the
gold-span advantage also observed on the instance-stratified split
(61.9\% to 65.3\%). The absence of
degradation relative to the primary split reduces concern that the main results
arise from memorizing shared base-passage scaffolds, although the splits differ
in size and class composition. Full results appear in
Appendix~\ref{app:base_disjoint}.

\subsection{The Gold-Span Oracle \& Predicted Spans}
We study the effect of localization quality by holding the encoder,
classifier, and training data fixed while varying the test-time span source (\Cref{tab:evidence_results}). We then examine how the predicted-to-gold gap varies across
inconsistency categories.

\paragraph{Per-class oracle decomposition.}
Table~\ref{tab:perclass_oracle} decomposes performance by category under the
fixed frozen-encoder pipeline, revealing that aggregate errors arise from
different sources. Referential shows the clearest sensitivity to localization:
its text-only F1 is 0.016 but rises to 0.540 when the correct spans are supplied.
Unit \& Measurement and Temporal also exhibit large predicted-to-gold gains.
By contrast, Factual and Logical remain difficult under gold evidence, reaching
F1 scores of only 0.454 and 0.493. These differences show that improving span
prediction is likely to benefit some categories substantially, whereas others
continue to require better semantic discrimination, background knowledge, or
reasoning even after localization is resolved.

\begin{table}[!htbp]
\centering
\caption{Per-class F1 under the fixed \emph{frozen} EmbeddingGemma + MLP
pipeline, isolating representation from span prediction. Text = no evidence;
Pred.\ = predicted spans (DeBERTa-v3-base); Gold = gold-span oracle.
$\Delta$ = Gold $-$ Pred. Rows ordered by $\Delta$.}
\label{tab:perclass_oracle}
\resizebox{\columnwidth}{!}{%
\begin{tabular}{l c c c c}
\toprule
\textbf{Category} & \textbf{Text} & \textbf{Pred.} & \textbf{Gold} & \textbf{$\Delta$} \\
\midrule
Unit \& Measurement  & 0.142 & 0.418 & 0.603 & 0.185 \\
Referential          & 0.016 & 0.360 & 0.540 & 0.180 \\
Temporal             & 0.207 & 0.472 & 0.611 & 0.139 \\
Factual              & 0.106 & 0.339 & 0.454 & 0.115 \\
Logical              & 0.244 & 0.378 & 0.493 & 0.115 \\
Numerical            & 0.263 & 0.624 & 0.715 & 0.091 \\
Specificity \& Scope & 0.038 & 0.515 & 0.601 & 0.086 \\
Pragmatic            & 0.082 & 0.459 & 0.540 & 0.081 \\
Theoretical          & 0.545 & 0.736 & 0.793 & 0.057 \\
Terminological       & 0.261 & 0.486 & 0.539 & 0.053 \\
Normative \& Policy  & 0.521 & 0.683 & 0.735 & 0.052 \\
\midrule
\textit{Accuracy}    & 0.271 & 0.510 & 0.609 & \\
\textit{Macro F1}    & 0.220 & 0.497 & 0.602 & \\
\bottomrule
\end{tabular}}
\end{table}

\paragraph{The oracle gain requires the right span, not any span.}
A potential concern is that the oracle gain is an artifact of the synthetic
construction: if the classifier benefits from being handed \emph{any} span
rather than specifically the \emph{correct} one, the gain would not reflect
genuine localization. We test this directly
by replacing the gold test spans with controls of the same form
(\Cref{tab:artifact_control}, Appendix~\ref{app:artifact_controls}). Random same-length spans reduce accuracy to 0.515, while high-TF--IDF
distractor spans reduce it to 0.502, both substantially below the 0.653
gold-span result. They also fall below the separately trained passage-only
classifier at 0.619, reflecting the fact that an evidence-augmented classifier
trained with relevant spans is actively misled when unrelated spans are supplied
at test time. The control therefore shows that the oracle improvement depends
on the content of the correct conflicting evidence rather than merely on adding
an auxiliary text segment.

\paragraph{Downstream performance generally tracks span quality.}
We examine the relationship between localization quality and classification
through a controlled mixing sweep (Appendix~\ref{app:fusion}). For each test
instance, each gold span is independently replaced by the model-predicted span
with probability $p \in \{0, 0.2, 0.4, 0.6, 0.8, 1.0\}$. This interpolates
between full gold evidence at $p=0$ (Token-F1 1.000; Macro-F1 0.645) and fully
predicted evidence at $p=1$ (Token-F1 0.679; Macro-F1 0.627). Evidence
Token-F1 decreases steadily as predicted spans replace gold spans, while
downstream Macro-F1 follows the same overall decline despite small
non-monotonic variations at intermediate points. This relationship supports
localization quality as a meaningful contributor to downstream performance,
rather than implying that it is the sole source of classification error.

\paragraph{The predicted-to-gold gap persists under end-to-end span-aware
training.}
Training the encoder jointly over the passage and evidence could, in principle,
make the classifier more robust to imperfect extracted spans. In practice, the
end-to-end model reaches 67.4\% accuracy with gold spans and 63.5\% with
predicted spans in the single-seed EmbeddingGemma-extractor condition. Predicted
evidence therefore improves over the 61.9\% passage-only baseline, but remains
approximately four points below the corresponding gold condition. The same qualitative pattern appears across the evaluated encoder regimes:
correct evidence yields the largest gain, predicted evidence realizes only part
of it, and the size of the gain depends on extractor quality. These results show
that span-prediction quality remains an important bottleneck even when the
classifier is trained jointly with evidence: the classifier can exploit
correctly localized claims, but the automatic extractor does not yet recover
the full benefit at test time. The remaining error under gold spans further
shows that improved localization must be paired with stronger type-level
reasoning and discrimination.

% \begin{table}[!htbp]
% \centering
% \caption{Oracle-artifact control (fine-tuned encoder, seed 42): gold test spans
% replaced by same-length random spans or high-TF-IDF distractor spans, training
% unchanged. Only the correct span recovers the oracle gain.}
% \label{tab:artifact_control}
% \begin{tabular}{l c c}
% \toprule
% \textbf{Test evidence} & \textbf{Acc.} & \textbf{Mac. F1} \\
% \midrule
% Gold spans \textit{(oracle)} & 0.653 & 0.645 \\
% Random same-length spans     & 0.515 & 0.502 \\
% TF-IDF distractor spans      & 0.502 & 0.481 \\
% \bottomrule
% \end{tabular}
% \end{table}

\paragraph{Span quality and role asymmetry.}
The selected extractor reaches 0.679 Token-F1 but only 0.371 exact match.
Localization is stronger for the conflicting claim
(\texttt{claim\_b}: 0.755 Token-F1) than for the anchoring claim
(\texttt{claim\_a}: 0.604), indicating that recovering the earlier statement
against which the conflict must be interpreted is the harder extraction step.
Full role-level results and fusion ablations appear in
Appendices~\ref{app:evidence_qa} and~\ref{app:fusion}.

\subsection{Confusion Analysis}
Per-class F1 identifies which categories are difficult but not how they fail. We
inspect the row-normalized confusion matrix of the fine-tuned
evidence-augmented pipeline (\Cref{tab:confusions}). Errors are structured rather than
uniform. Two patterns stand out. First, Logical acts as a broad attractor, absorbing errors from Pragmatic (20.1\% of its rows), Specificity
\& Scope (14.1\%), Referential (15.6\%), Theoretical (11.6\%), and Normative \&
Policy (7.9\%). Second, Factual and Temporal are mutually confused: Factual is
most often mistaken for Temporal (20.4\%) and Temporal for Factual (13.8\%),
suggesting that some factual conflicts surface through reporting-period or
event-sequence cues. Referential is unstable, splitting its errors between
Logical (15.6\%) and Factual (14.7\%). These patterns extend the oracle analysis:
residual errors reflect both local overlap between neighboring labels and a
broader collapse of difficult cases into reasoning-oriented categories.
Some of this overlap may also reflect ambiguity in the taxonomy itself rather
than model error alone, particularly for closely related categories such as
Factual, Temporal, and Logical. Distinguishing taxonomy ambiguity from residual model discrimination error would benefit from human validation.

\begin{table}[!htbp]
\centering
\caption{Largest off-diagonal confusions of the fine-tuned evidence-augmented
pipeline, by row-normalized percentage of the gold class. Rows sorted by Row \%.}
\label{tab:confusions}
\resizebox{\columnwidth}{!}{%
\begin{tabular}{l l c c}
\toprule
\textbf{Gold} & \textbf{Predicted} & \textbf{Count} & \textbf{Row \%} \\
\midrule
Factual              & Temporal       & 29 & 20.4 \\
Pragmatic            & Logical        & 29 & 20.1 \\
Referential          & Logical        & 17 & 15.6 \\
Referential          & Factual        & 16 & 14.7 \\
Specificity \& Scope & Logical        & 21 & 14.1 \\
Temporal             & Factual        & 23 & 13.8 \\
Pragmatic            & Temporal       & 19 & 13.2 \\
Theoretical          & Logical        & 21 & 11.6 \\
Logical              & Normative \& Policy & 19 & 7.9 \\
Normative \& Policy  & Logical        & 18 & 7.9 \\
\bottomrule
\end{tabular}}
\end{table}

\section{Conclusion}
We studied fine-grained inconsistency classification in financial disclosures
under a shared 11-class evaluation protocol. Supervised fine-tuning produced
large gains over frozen representations  across all three encoder backbones tested, and a
fine-tuned 300M encoder performed competitively with substantially larger
prompted and LoRA-adapted models at much lower computational cost. Because the
systems differ in architecture, supervision, and optimization, this should be
interpreted as a practical efficiency result rather than a controlled conclusion
about model scale.

The evidence experiments help distinguish localization-related error from
residual type-discrimination error. Gold spans consistently improve classification under both the instance-stratified and base-passage-disjoint evaluations, while predicted spans recover only part of this benefit. Referential, Unit \&
Measurement, and Temporal inconsistencies are especially sensitive to
localization quality, whereas Factual and Logical inconsistencies remain
difficult even with correct evidence. These findings indicate that progress
requires both stronger span extraction and better discrimination among related
inconsistency types. Evaluation on longer and naturally occurring financial
documents remains an important next step.

\paragraph{Acknowledgement.} We would like to thank Chi Heem Wong for supporting this work by providing the snapshot of SBID-FD data.

\paragraph{Reproducibility.}
We provide the code and resources required to reproduce the experiments in this
paper.\footnote{\url{https://github.com/amankumarhal/financial_inconsistency}}
The repository includes the classifier and extractor training scripts,
evidence-augmented and oracle pipelines, evaluation and analysis scripts, and
the exact train--test split used in this study. All hyperparameters, data splits,
and random seeds are documented in the appendices.

\section*{Limitations}
The study uses short synthetic passages with one localized inconsistency.
Naturally occurring financial errors may be sparse, span distant sections,
involve tables or external filings, or arise through drafting processes not
represented by the generation pipeline, so the results characterize a
controlled diagnostic setting rather than complete filings.

The task conditions on an inconsistency already being detected and does not
measure false positives, false negatives, or upstream error propagation.

The evidence analysis assumes two localized verbatim spans and paired span
supervision during training. The gold condition represents a best-case
localization setting under this pipeline, while the 
automatic extractor (0.679 Token-F1, 0.371 exact match) is not a
deployment-ready localization system.

The base-passage-disjoint evaluation reduces concern about scaffold overlap
but not broader generation artifacts. In particular, Qwen3.5 models appear
in both the generator pool and the evaluated systems, and GPT-5-mini shares
a family with GPT-5.4, so familiarity with generator outputs could favor
related models. Extractor, end-to-end, artifact-control, and several
large-model results are single-seed, so variance estimates remain
incomplete.

\paragraph{Societal impact.}
The evaluated systems should be treated only as human-reviewed triage aids.
Missed, misclassified, or incorrectly localized inconsistencies could create
false confidence in consequential financial disclosures. Predictions should
therefore be accompanied by supporting evidence and verified by qualified human
reviewers.

% \paragraph{Societal impact.}
% Fine-grained inconsistency classification could assist auditors and financial
% analysts by triaging candidate conflicts and directing different inconsistency
% types toward appropriate downstream checks. However, false negatives,
% misclassification, and incorrectly localized evidence could create unwarranted
% confidence in consequential disclosures. The evaluated systems should therefore
% be viewed as diagnostic or triage aids, not as automated verifiers or substitutes
% for professional review. In any future use, predictions should be presented
% with their supporting evidence and uncertainty, and consequential decisions
% should remain subject to human expert verification.

% \section*{Acknowledgments}

% The authors used AI writing assistants (Claude and ChatGPT) for coding support
% and paper editing. All experimental results, analyses, and scientific claims are
% the authors' own.

\bibliography{latex/custom}

\appendix

\section{Dataset Examples}
\label{app:examples}
To make the task concrete, we show one dataset instance illustrating the paired
evidence-span structure used in the localization analysis: two conflicting clauses placed
inside a longer passage, each marked by a gold span, with the passage typed into
one of 11 inconsistency categories. \Cref{tab:class_counts} gives the per-class
counts for both splits.

\begin{table}[!htbp]
\centering
\caption{Per-class instance counts in the train and test splits (stratified
70--30).}
\label{tab:class_counts}
\begin{tabular}{l c c}
\toprule
\textbf{Inconsistency type} & \textbf{Train} & \textbf{Test} \\
\midrule
Logical              & 564 & 241 \\
Normative \& Policy  & 531 & 228 \\
Theoretical          & 423 & 181 \\
Numerical            & 419 & 180 \\
Temporal             & 390 & 167 \\
Specificity \& Scope & 347 & 149 \\
Pragmatic            & 337 & 144 \\
Factual              & 330 & 142 \\
Terminological       & 289 & 124 \\
Unit \& Measurement  & 273 & 117 \\
Referential          & 255 & 109 \\
\midrule
\textit{Total}       & 4{,}158 & 1{,}782 \\
\bottomrule
\end{tabular}
\end{table}

\paragraph{Example of 11-way typing with paired evidence spans.}
\emph{Label:} Logical inconsistency.
\emph{Passage:} ``Projected earnings per share of \$2.40 assumes stable diesel
fuel pricing across our regional distribution network. Freight rate increases
exceeding 8\% year-over-year would compress operating margins below our
conservative targets. Operating margins are structurally insulated from freight
cost volatility. Fleet utilization rates are forecast to reach 85\% by Q4,
contingent on hiring 500 new drivers. Fuel hedging strategies cover only 60\% of our anticipated consumption volume for the next two quarters. Macro-economic headwinds could invalidate these operational assumptions entirely.''
\emph{Gold spans:} \textsc{claim\_a} = ``Freight rate increases exceeding 8\%
year-over-year would compress operating margins below our conservative
targets''; \textsc{claim\_b} = ``Operating margins are structurally insulated
from freight cost volatility''.
\emph{Why inconsistent:} the passage first states that higher freight rates
would compress margins, then asserts margins are insulated from freight
volatility; the two claims cannot both hold. The task is to predict the type
(here Logical) given the passage; the evidence-augmented pipeline additionally
supplies the two spans.

\paragraph{Length statistics.}
\Cref{tab:length_stats} reports length statistics for the base passages and their
inconsistent variants over the 5,940 instances. Token counts use the Qwen3.5
tokenizer. The train and test splits are nearly identical on these statistics
(the split is stratified), so we report the pooled figures.

\begin{table}[!htbp]
\centering
\caption{Length statistics for base passages and inconsistent variants over the
5,940 instances. Token counts use the Qwen3.5 tokenizer. Variants are slightly
longer, reflecting the information added to introduce a conflict.}
\label{tab:length_stats}
\resizebox{\columnwidth}{!}{%
\begin{tabular}{lcccc}
\toprule
Metric & Mean & S.D. & Min & Max \\
\midrule
\multicolumn{5}{l}{\emph{Base passages}} \\
Characters            & 574.9 & 125.4 & 73 & 1{,}099 \\
Tokens (Qwen3.5)      & 100.9 & 23.3  & 12 & 200 \\
Sentences             & 5.2   & 1.5   & 1  & 12 \\
\midrule
\multicolumn{5}{l}{\emph{Inconsistent variants}} \\
Characters            & 625.8 & 142.2 & 97 & 1{,}176 \\
Tokens (Qwen3.5)      & 110.5 & 26.5  & 19 & 236 \\
Sentences             & 5.6   & 1.7   & 1  & 14 \\
\bottomrule
\end{tabular}}
\end{table}

\paragraph{Surface-form overlap.}
To quantify how localized the perturbations are, we measure character-level edit
(Levenshtein) distance and longest common contiguous substring (LCS) between each
base passage and its inconsistent variant (\Cref{tab:surface_overlap}). The
median edit distance is 72 characters and the median LCS covers about 73\% of the
shorter passage, confirming that inconsistencies are introduced through localized
edits or insertions rather than wholesale rewrites.

\begin{table}[!htbp]
\centering
\caption{Surface-form overlap between base passages and their inconsistent
variants over the 5,940 instances. Levenshtein distance is character-level; LCS
is the longest common contiguous substring. Normalized LCS divides the LCS length
by the shorter (min) or longer (max) passage length.}
\label{tab:surface_overlap}
\resizebox{\columnwidth}{!}{%
\begin{tabular}{lrrrr}
\toprule
\textbf{Metric} & \textbf{Mean} & \textbf{Median} & \textbf{P5} & \textbf{P95} \\
\midrule
Levenshtein distance & 81.8  & 72    & 9   & 198 \\
LCS length (chars)   & 420.1 & 404   & 213 & 682 \\
LCS / min length     & 0.742 & 0.734 & --  & 1.000 \\
LCS / max length     & 0.661 & 0.655 & --  & -- \\
\bottomrule
\end{tabular}}
\end{table}

\section{Base-Passage-Disjoint Evaluation}
\label{app:base_disjoint}

The primary split is stratified at the instance level and may place variants
derived from the same base passage in both training and test. For the base-passage-disjoint evaluation, we group examples by their
underlying base passage and assign each group entirely to one partition.
The resulting test set contains 1,813 examples.
\Cref{tab:base_disjoint} compares the primary and base-disjoint splits
across the three evidence conditions.

\begin{table}[!htbp]
\centering
\caption{Comparison between the primary instance-stratified split and the
base-passage-disjoint split.}
\label{tab:base_disjoint}
\resizebox{\columnwidth}{!}{%
\begin{tabular}{lcccc}
\toprule
& \multicolumn{2}{c}{\textbf{Primary}} &
  \multicolumn{2}{c}{\textbf{Base-disjoint}} \\
\cmidrule(lr){2-3}
\cmidrule(lr){4-5}
\textbf{Method} &
\textbf{Acc.} & \textbf{Mac. F1} &
\textbf{Acc.} & \textbf{Mac. F1} \\
\midrule
Passage only
& 0.619 & 0.614 & 0.630 & 0.633 \\

Predicted spans
& 0.633 & 0.627 & 0.630 & 0.631 \\

Gold spans
& 0.653 & 0.645 & 0.650 & 0.653 \\
\bottomrule
\end{tabular}}
\end{table}

All three conditions use the fine-tuned EmbeddingGemma encoder (trained on
the respective split) frozen as a feature extractor for the evidence-augmented
runs; the passage-only row uses the same encoder without an evidence branch.
On the base-disjoint test set, the fine-tuned passage-only and
predicted-span models achieve the same accuracy (1,142/1,813 correct) but
produce different labels on 119 instances. Among these 119 disagreements,
predicted spans correct 30 passage-only errors and overturn 30 passage-only
correct predictions. Gold evidence yields 1,179 correct predictions, a net
gain of 37 over passage-only classification. Fine-tuned passage-only
accuracy is slightly higher on the base-disjoint split than on the primary
split (63.0\% vs.\ 61.9\%), arguing against scaffold memorization. The gold-oracle benefit persists under both splits, whereas the automatic
predicted-span gain on the primary split (+1.4 pp) does not yield a net
accuracy improvement here.

\section{Evidence-Span QA: Implementation Details}
\label{app:evidence_qa}

\subsection*{Task Formulation}
We model evidence-span prediction as an extractive QA problem over each 
sample's inconsistent passage. Each instance contains a gold evidence list 
with items of the form \texttt{\{"quote": "...", "role": "claim\_a"|"claim\_b"\}}, 
converted into up to two QA training examples:
\begin{itemize}
    \item \textit{Question:} ``Extract the evidence span for claim A 
    (verbatim substring).'' \textit{Answer:} gold quote with 
    \texttt{role="claim\_a"}
    \item \textit{Question:} ``Extract the evidence span for claim B 
    (verbatim substring).'' \textit{Answer:} gold quote with 
    \texttt{role="claim\_b"}
\end{itemize}
The training split expands from 4,158 rows to 8,076 QA examples; the test
split expands from 1,782 rows to 3,481 examples. In both splits, each row
yields up to two QA examples (one per role); examples whose gold quote cannot
be aligned to \texttt{inconsistent\_text} (exact match, then
whitespace-normalized match) are skipped. This excludes 240 training QA
examples (104 \texttt{claim\_a}, 136 \texttt{claim\_b}) and 83 test QA
examples (34 \texttt{claim\_a}, 49 \texttt{claim\_b}). The remaining rows
may contribute one or both roles depending on which quotes align.

\subsection*{Model and Training}
We fine-tune five encoder backbones with an extractive QA head
(\texttt{max\_length=384}, \texttt{doc\_stride=128}):
\texttt{distilbert-base-uncased} \citep{sanh2019distilbert},
\texttt{roberta-base} \citep{liu2019roberta},
\texttt{bert-base-uncased} \citep{devlin2019bert},
\texttt{deberta-v3-base} \citep{he2023debertav3}, and the EmbeddingGemma-300M
backbone. For the last, EmbeddingGemma is
released as an embedding model with a mean-pooling head; we discard that head
and attach a token-level start/end span head to the underlying
\texttt{Gemma3TextModel} transformer, training it with the identical QA protocol.
Each backbone is trained on the training split with a 10\% passage-level
validation slice held out for checkpoint selection; the test split is not used
during training. We select the EmbeddingGemma QA extractor for the primary predicted-span
experiments because it achieves the highest span quality
(\Cref{tab:qa_backbone_sweep}); training was stable in fp32 with no divergence.

\subsection*{Inference and Decoding}
At test time, the model is run twice per instance, once per role. Candidate 
spans are scored by:
\begin{equation}
    \text{score}(s, e) = \text{start\_logit}[s] + \text{end\_logit}[e]
\end{equation}
searching over the top-25 start and end positions across all overflow windows,
keeping the highest-scoring span of at most 80 tokens. \Cref{tab:qa_role_results} reports exact-match and Token-F1 by role for
the selected extractor; localization is stronger for \texttt{claim\_b}
than for \texttt{claim\_a}, consistent with the role asymmetry discussed
in the main text.

\begin{table}[!htbp]
\centering
\caption{QA backbone sweep results (seed 42; 10\% train validation,
test held out until inference). EmbeddingGemma is selected for the primary
predicted-span experiments.} 
\label{tab:qa_backbone_sweep}
\begin{tabular}{l c c}
\toprule
\textbf{Backbone} & \textbf{EM} & \textbf{Token-F1} \\
\midrule
\texttt{EmbeddingGemma-QA}    & \textbf{0.3710} & \textbf{0.6790} \\
\texttt{FinBERT (ProsusAI)}   & 0.3238 & 0.6130 \\
\texttt{deberta-v3-base}      & 0.3030 & 0.6110 \\
\texttt{distilbert-base-unc.} & 0.2784 & 0.5740 \\
\texttt{bert-base-uncased}    & 0.2611 & 0.5464 \\
\texttt{roberta-base}         & 0.1864 & 0.4529 \\
\bottomrule
\end{tabular}
\end{table}

\begin{table}[!htbp]
\centering
\caption{Evidence-span QA results by role for the selected extractor
(EmbeddingGemma QA, seed 42; trained with 10\% train-only validation,
test held out until inference).}
\label{tab:qa_role_results}
\begin{tabular}{l c c c}
\toprule
\textbf{Role} & \textbf{EM} & \textbf{Token-F1} & \textbf{n} \\
\midrule
\texttt{claim\_a} & 0.344 & 0.6040 & 1,748 \\
\texttt{claim\_b} & 0.399 & 0.7550 & 1,733 \\
\midrule
Overall           & 0.3710 & 0.6790 & 3,481 \\
\bottomrule
\end{tabular}
\end{table}

\section{EmbeddingGemma as a Span Extractor}
\label{app:gemma_extractor}
Our selected extractor attaches a QA head to the EmbeddingGemma backbone. We
summarize the experiments supporting this choice.

\paragraph{Span quality.} Under the identical QA protocol (seed 42, 10\%
passage validation, \texttt{max\_length} 384, \texttt{doc\_stride} 128), the
EmbeddingGemma QA extractor attains overall Token-F1 0.679 (EM 0.371), versus
0.611 (EM 0.303) for DeBERTa-v3-base (\Cref{tab:qa_backbone_sweep}). Training
was stable in fp32 with no divergence. The released EmbeddingGemma checkpoint is
an embedding model with a pooling head; we discard that head and train the
underlying transformer with a token-level span head, confirming that the
backbone supports extractive span prediction and, here, does so better than a
standard QA encoder.

\paragraph{Downstream effect.}
Swapping only the test-time predicted spans in the primary
evidence-augmented pipeline (fine-tuned EmbeddingGemma classifier, pooled MLP,
seed 42) changes accuracy from 0.621 with DeBERTa spans to 0.633 with
EmbeddingGemma spans, and Macro-F1 from 0.614 to 0.627. The corresponding gold-span result is 0.653 accuracy and 0.645 Macro-F1.
Higher span quality improves downstream classification, though a gap to the
matched gold condition remains.

\paragraph{Joint multitask training underperforms the two-model pipeline.}
We also train a single Gemma backbone jointly with a span head and an 11-way
classification head, using its own predicted spans for classification. Its span
quality remains high at 0.675 Token-F1, but classification reaches only 0.604
accuracy and 0.586 Macro-F1, compared with 0.633 and 0.627 for the separate
extractor and classifier. We therefore retain the two-model pipeline in the
primary experiments.

\section{Classifier Training Details}
\label{app:lora_traditional}
All MLP heads use hidden layers [512, 256], dropout 0.2, lr $=10^{-3}$, weight
decay $=10^{-4}$, batch size 256, 25 epochs. Frozen-encoder features are
precomputed. The fine-tuned EmbeddingGemma encoder is trained end-to-end on
passage-only 11-way classification (5 epochs, lr $=2\times10^{-5}$, batch size
8); for the evidence-augmented and oracle conditions, this fine-tuned encoder is
then used as a frozen feature extractor for both passage and span embeddings,
and only the MLP head is trained. This is not full end-to-end span-aware
fine-tuning; the encoder is adapted on passages and reused as an embedder.
Fine-tuned DeBERTa-v3-base passage classification uses 5 epochs, lr
$=6\times10^{-6}$, warmup 0.1. The end-to-end span-aware variant instead tokenizes a single string formed by
appending gold \texttt{claim\_a} and \texttt{claim\_b} verbatim to the passage
(\texttt{passage\,|\,evidence: $\langle$claim\_a$\rangle$ $\langle$claim\_b$\rangle$}),
fine-tunes the full EmbeddingGemma encoder jointly with a \emph{linear}
11-way head (5 epochs, lr $=2\times10^{-5}$, batch size 8, seed~42), and at
test replaces the appended spans with QA-predicted quotes while keeping the
same input template.

\paragraph{Interpreting the frozen evidence results.}
In the frozen-encoder setting, adding predicted spans raises accuracy from
27.1\% for passage-only classification to 51.0\%. This large difference should
not be interpreted as the isolated effect of accurate localization, because the
frozen passage representation is poorly aligned with the 11-class task and the
additional span representation supplies substantial task-relevant information.
The gold-span result reaches 60.9\%, versus 65.3\% with the passage-adapted
encoder. These are oracle results for their respective pipelines, not upper
bounds for other fusion methods or end-to-end architectures.

\paragraph{Compute.} Experiments were run on a node with four NVIDIA
RTX~6000 Ada Generation GPUs, though most use only
a single GPU; multiple GPUs were used only for the largest model (the 27B LoRA
run). Approximate wall-clock training times (recovered from run logs) illustrate
the cost profile. Frozen-encoder runs are dominated by a single embedding pass
and take seconds to a few minutes. Fine-tuning the EmbeddingGemma passage
classifier takes about four minutes per seed on a single GPU; the
evidence-augmented head trains in well under a minute on top of cached features,
and the end-to-end span-aware variant takes about twenty minutes. The selected
EmbeddingGemma QA extractor trains in about two and a half minutes. By contrast, LoRA generative baselines are substantially more expensive:
Qwen3.5-9B takes approx. fifty minutes per seed and Qwen3.5-27B about
two and a half hours, while achieving accuracy close to that of the
approximately four-minute encoder fine-tuning run. Prompted inference over the test set ranges from a few minutes
for the smaller models to approximately one hour
for the largest. These figures are approximate and
recovered post hoc; some runs lacked reliable timing.

\section{Distractor-Span Controls}
\label{app:artifact_controls}

We test whether the gold-span gain arises from the correct conflicting evidence
or merely from supplying additional text. The evidence-augmented classifier is
trained with gold spans, while the test spans are replaced by either random
same-length spans or high-TF--IDF distractor spans from the same passage.

\begin{table}[!htbp]
\centering
\caption{Distractor-span controls using the fine-tuned encoder at seed 42.
Training is unchanged; only the test-time evidence source is varied.}
\label{tab:artifact_control}
\begin{tabular}{lcc}
\toprule
\textbf{Test evidence} & \textbf{Acc.} & \textbf{Mac. F1} \\
\midrule
Gold spans \textit{(oracle)} & 0.653 & 0.645 \\
Random same-length spans     & 0.515 & 0.502 \\
TF--IDF distractor spans     & 0.502 & 0.481 \\
\bottomrule
\end{tabular}
\end{table}

Both distractor conditions perform substantially below the gold-span condition
and below the separately trained passage-only model at 0.619 accuracy. Because
the classifier was trained with relevant spans, unrelated spans create a
test-time distribution shift and actively mislead the evidence branch. The
control therefore shows that the oracle gain depends on the content of the
correct conflicting evidence rather than on adding arbitrary auxiliary text.

\begin{table}[!htbp]
\centering
\caption{Feature-fusion ablation using the fine-tuned EmbeddingGemma encoder
as a frozen feature extractor and DeBERTa-predicted spans (seed 42). Accuracy
is reported for predicted and gold evidence. These runs use the fusion-ablation
implementation and are reported separately from the primary extractor
comparison in \Cref{tab:evidence_results}.}
\label{tab:fusion}
\begin{tabular}{l c c}
\toprule
\textbf{Fusion} & \textbf{Pred.} & \textbf{Gold} \\
\midrule
Mean-pool           & 0.621 & 0.653 \\
Two-slot concat     & 0.619 & 0.654 \\
Max-pool            & 0.619 & 0.652 \\
Cross-encoder       & 0.607 & 0.671 \\
\bottomrule
\end{tabular}
\end{table} 

\section{Feature-Fusion Ablation and Span-Quality Sweep}
\label{app:fusion}
We ablate how passage and span representations are combined in the
fine-tuned-encoder feature pipeline (seed 42), reporting both predicted-span and
gold-span conditions (\Cref{tab:fusion}). Mean pooling, two-slot concatenation,
and max pooling produce similar results and preserve a predicted-to-gold gap of
approximately three points. A single-sequence cross-encoder raises the gold-span
accuracy to 0.671 but does not improve the predicted-span condition, which
reaches 0.607. These results indicate that more expressive fusion better exploits accurate
evidence but does not compensate for extraction errors. Localization quality
therefore remains a constraint alongside residual type-discrimination errors
under gold evidence.

\paragraph{Span-quality mixing sweep.}
To verify that downstream accuracy tracks span quality, we vary the fraction
of gold test spans via per-role Bernoulli substitution with probability
$p$ (probability of replacing a gold span with the
EmbeddingGemma-QA-predicted span).
\Cref{tab:span_sweep} reports the results on the fine-tuned EmbeddingGemma
pipeline. Span Token-F1 falls steadily as $p$ increases; downstream Macro-F1
tracks the endpoints and the overall trend, though it is not strictly monotone
at intermediate values.

\begin{table}[!htbp]
\centering
\caption{Span-quality mixing sweep on the fine-tuned EmbeddingGemma pipeline
(seed 42). $p$ = probability of substituting the EmbeddingGemma QA-predicted
span for the gold span per role. $p=0$: full gold spans; $p=1$: full predicted
spans. Downstream Macro-F1 at $p=1$ matches the single-seed result in
\Cref{tab:evidence_results} (Macro-F1 0.627).}
\label{tab:span_sweep}
\begin{tabular}{c c c}
\toprule
$p$ & \textbf{Avg.\ Token-F1} & \textbf{Macro-F1} \\
\midrule
0.0 & 1.000 & 0.645 \\
0.2 & 0.936 & 0.644 \\
0.4 & 0.872 & 0.635 \\
0.6 & 0.807 & 0.637 \\
0.8 & 0.744 & 0.634 \\
1.0 & 0.679 & 0.627 \\
\bottomrule
\end{tabular}
\end{table}

\section{LoRA Fine-Tuning Details}
\label{app:lora}
We fine-tune \texttt{Qwen/Qwen3.5-9B} under 4-bit NF4 (Normal Float 4) quantization 
(\texttt{bitsandbytes}, \citealp{bitsandbytes2026repo}) with double quantization and \texttt{bf16} compute 
dtype. LoRA adapters are applied to attention projections 
(\texttt{q\_proj}, \texttt{k\_proj}, \texttt{v\_proj}, \texttt{o\_proj}), 
with rank $r = 16$, scaling $\alpha = 32$, and dropout $0.05$. Training 
uses 1 epoch, \texttt{paged\_adamw\_8bit}, learning rate $2\times10^{-4}$ 
with warmup ratio $0.03$, effective batch size 16, and maximum sequence 
length 1,024 tokens. At inference, we use deterministic decoding with 
\texttt{max\_new\_tokens=32} and parse predictions via exact-line matching 
with a substring fallback. The \texttt{Qwen/Qwen3.5-27B} LoRA model uses the identical configuration
(same rank, schedule, prompt, and decoding), changing only the base checkpoint
and providing a more controlled within-family comparison of model size; it is
run for a single seed.

\begin{table}[!htbp]
\centering
\caption{Per-class F1 for fine-tuned FinBERT and EmbeddingGemma passage
classifiers and the EmbeddingGemma evidence-augmented pipeline (seed 42).
FinBERT and EmbGemma use the fine-tuned encoder + MLP head; DeB.\ pred. and
EmbG.\ pred. denote predicted spans from the DeBERTa and EmbeddingGemma
extractors, respectively; Gold is the gold-span oracle.}
\label{tab:perclass_finetuned}

\setlength{\tabcolsep}{2.5pt}
\renewcommand{\arraystretch}{1.02}

\resizebox{\columnwidth}{!}{%
\begin{tabular}{@{}lccccc@{}}
\toprule
\textbf{Category}
& \textbf{FinBERT}
& \textbf{EmbG.}
& \textbf{DeB. pred.}
& \textbf{EmbG. pred.}
& \textbf{Gold} \\
\midrule
Factual              & 0.323 & 0.403 & 0.404 & 0.431 & 0.449 \\
Logical              & 0.397 & 0.460 & 0.473 & 0.489 & 0.527 \\
Normative \& Policy  & 0.753 & 0.788 & 0.792 & 0.802 & 0.804 \\
Numerical            & 0.725 & 0.731 & 0.760 & 0.751 & 0.767 \\
Pragmatic            & 0.523 & 0.551 & 0.534 & 0.555 & 0.563 \\
Referential          & 0.271 & 0.434 & 0.442 & 0.478 & 0.500 \\
Specificity \& Scope & 0.582 & 0.609 & 0.625 & 0.641 & 0.672 \\
Temporal             & 0.567 & 0.541 & 0.559 & 0.571 & 0.607 \\
Terminological       & 0.673 & 0.648 & 0.657 & 0.654 & 0.652 \\
Theoretical          & 0.823 & 0.781 & 0.773 & 0.786 & 0.798 \\
Unit \& Measure.     & 0.627 & 0.700 & 0.730 & 0.737 & 0.760 \\
\midrule
\textit{Accuracy}    & 0.579 & 0.612 & 0.621 & 0.633 & 0.653 \\
\textit{Macro F1}    & 0.570 & 0.604 & 0.614 & 0.627 & 0.645 \\
\bottomrule
\end{tabular}}
\end{table}

\section{Per-Class F1 Scores: Frozen-Encoder Methods}
\label{app:perclass_traditional}
\Cref{tab:perclass_traditional_consolidated} reports per-class F1 for all frozen text-only backbones, including
FinBERT, and for the frozen EmbeddingGemma pipeline with predicted and gold
spans, the configuration used for the localization decomposition in \Cref{tab:perclass_oracle}.

\begin{table*}[t!]
\centering
\caption{Per-class F1 scores for frozen-encoder MLP classifiers (1,782 test
examples, seed 42). EmbGemma = EmbeddingGemma-300M; MiniLM = all-MiniLM-L6-v2;
QwenEmb = Qwen3-Embedding; FinBERT = ProsusAI/finbert;
pred.\ = predicted evidence spans (DeBERTa-v3-base);
gold = gold-span oracle.}
\label{tab:perclass_traditional_consolidated}
\resizebox{\linewidth}{!}{%
\begin{tabular}{l c c c c c c c c c}
\toprule
\textbf{Category} & \textbf{n} & \textbf{MiniLM} & \textbf{QwenEmb-0.6B}
    & \textbf{QwenEmb-8B} & \textbf{DeBERTa} & \textbf{EmbGemma}
    & \textbf{FinBERT} & \textbf{EmbGemma (pred.)} & \textbf{EmbGemma (gold)} \\
\midrule
Factual                  & 142 & 0.095 & 0.084 & 0.133 & 0.124 & 0.106 & 0.064 & 0.339 & 0.454 \\
Logical                  & 241 & 0.131 & 0.180 & 0.184 & 0.244 & 0.244 & 0.187 & 0.378 & 0.493 \\
Normative \& Policy      & 228 & 0.335 & 0.428 & 0.507 & 0.306 & 0.521 & 0.426 & 0.683 & 0.735 \\
Numerical                & 180 & 0.151 & 0.239 & 0.224 & 0.473 & 0.263 & 0.352 & 0.624 & 0.715 \\
Pragmatic                & 144 & 0.127 & 0.120 & 0.194 & 0.000 & 0.082 & 0.130 & 0.459 & 0.540 \\
Referential              & 109 & 0.031 & 0.047 & 0.043 & 0.000 & 0.016 & 0.000 & 0.360 & 0.540 \\
Specificity \& Scope     & 149 & 0.082 & 0.022 & 0.071 & 0.000 & 0.038 & 0.186 & 0.515 & 0.601 \\
Temporal                 & 167 & 0.178 & 0.109 & 0.226 & 0.241 & 0.207 & 0.370 & 0.472 & 0.611 \\
Terminological           & 124 & 0.135 & 0.110 & 0.094 & 0.000 & 0.261 & 0.127 & 0.486 & 0.539 \\
Theoretical              & 181 & 0.391 & 0.396 & 0.467 & 0.235 & 0.545 & 0.446 & 0.736 & 0.793 \\
Unit \& Measurement      & 117 & 0.069 & 0.062 & 0.078 & 0.033 & 0.142 & 0.016 & 0.418 & 0.603 \\
\midrule
Accuracy                 &     & 0.189 & 0.204 & 0.233 & 0.234 & 0.271 & 0.276 & 0.510 & 0.609 \\
Macro F1                 &     & 0.157 & 0.163 & 0.202 & 0.151 & 0.220 & 0.209 & 0.497 & 0.602 \\
Weighted F1              &     & 0.173 & 0.187 & 0.227 & 0.178 & 0.248 & 0.236 & 0.511 & 0.611 \\
\bottomrule
\end{tabular}}
\end{table*}

\section{Per-Class F1 Scores: Fine-Tuned Encoder}
\label{app:perclass_finetuned}
\Cref{tab:perclass_finetuned} reports per-class F1 for fine-tuned FinBERT
and EmbeddingGemma passage classifiers and the EmbeddingGemma evidence pipeline
with DeBERTa-predicted, EmbeddingGemma-predicted, and gold spans. EmbeddingGemma
generally outperforms FinBERT at passage level, while its extractor improves
most categories over DeBERTa; nearly all remain below the gold oracle.

\section{Per-Class F1 Scores: Generative Models}
\label{app:perclass_generative}
\Cref{tab:perclass_generative_consolidated} reports per-class F1 for the generative
configurations, ordered by accuracy. Prompted Qwen3.5-9B is uneven across classes,
including weak Theoretical performance \& over-prediction of Logical,
suggesting label bias without task-specific adaptation. GPT-5.4 performs
strongly on Numerical and Normative \& Policy but remains weaker on Factual
and Pragmatic. LoRA adaptation makes Qwen3.5-9B substantially more balanced,
with strong Theoretical and Unit \& Measurement performance, although Factual
remains difficult.

\begin{table*}[t!]
\centering
\caption{Per-class F1 scores for generative models (1,782 test examples). 
All prompted models use label definitions and illustrative examples. Models 
ordered by overall accuracy. LoRA = Qwen3.5-9B QLoRA SFT with the same
definitions+examples prompt as Table~\ref{tab:generative_results}.
All columns are single-seed (seed~42) for per-class detail; the LoRA column
is seed~42 of the three seeds (1, 2, 42) averaged in
\Cref{tab:generative_results} (0.617 acc, 0.597 macro for this seed;
three-seed means $0.615 \pm 0.002$ and $0.596 \pm 0.002$).}
\label{tab:perclass_generative_consolidated}
\resizebox{\linewidth}{!}{%
\begin{tabular}{l c c c c c c c c c c}
\toprule
\textbf{Category} & \textbf{n} 
& \textbf{Qwen3.5-4B} 
& \textbf{Qwen3.5-9B} 
& \textbf{Qwen3.5-35B-A3B} 
& \textbf{Gemma-4-31B} 
& \textbf{Gemma-4-E4B} 
& \textbf{Gemma-4-26B} 
& \textbf{Qwen3.5-27B} 
& \textbf{GPT-5.4} 
& \textbf{Qwen3.5-9B + LoRA} \\
\midrule
Factual             & 142 & 0.000 & 0.138 & 0.207 & 0.153 & 0.152 & 0.165 
    & 0.171 & 0.258 & 0.241 \\
Logical             & 241 & 0.222 & 0.343 & 0.342 & 0.346 & 0.414 & 0.384 
    & 0.447 & 0.566 & 0.517 \\
Normative \& Policy & 228 & 0.547 & 0.537 & 0.651 & 0.622 & 0.655 & 0.621 
    & 0.687 & 0.745 & 0.739 \\
Numerical           & 180 & 0.369 & 0.269 & 0.580 & 0.683 & 0.552 & 0.683 
    & 0.629 & 0.796 & 0.729 \\
Pragmatic           & 144 & 0.098 & 0.123 & 0.221 & 0.133 & 0.106 & 0.093 
    & 0.321 & 0.305 & 0.578 \\
Referential         & 109 & 0.227 & 0.229 & 0.372 & 0.412 & 0.380 & 0.331 
    & 0.478 & 0.556 & 0.436 \\
Specificity \& Scope & 149 & 0.283 & 0.157 & 0.244 & 0.274 & 0.536 & 0.449 
    & 0.502 & 0.703 & 0.569 \\
Temporal            & 167 & 0.249 & 0.470 & 0.531 & 0.529 & 0.523 & 0.539 
    & 0.584 & 0.661 & 0.625 \\
Terminological      & 124 & 0.447 & 0.410 & 0.158 & 0.400 & 0.386 & 0.398 
    & 0.519 & 0.607 & 0.608 \\
Theoretical         & 181 & 0.064 & 0.085 & 0.162 & 0.290 & 0.276 & 0.593 
    & 0.339 & 0.594 & 0.778 \\
Unit \& Measurement & 117 & 0.180 & 0.267 & 0.097 & 0.185 & 0.390 & 0.265 
    & 0.514 & 0.694 & 0.746 \\
\midrule
Accuracy            &     & 0.274 & 0.330 & 0.389 & 0.404 & 0.409 & 0.439 
    & 0.487 & 0.613 & 0.617 \\
Macro F1            &     & 0.244 & 0.275 & 0.324 & 0.366 & 0.397 & 0.411 
    & 0.472 & 0.589 & 0.597 \\
Weighted F1         &     & 0.256 & 0.289 & 0.349 & 0.385 & 0.414 & 0.434 
    & 0.482 & 0.600 & 0.607 \\
\bottomrule
\end{tabular}}
\end{table*}

% \section{Classification Prompt}
% \label{app:prompt}
% \begin{tcolorbox}[
%     colback=gray!5,
%     colframe=gray!50,
%     title={\small\textbf{Classification Prompt}},
%     fontupper=\small\ttfamily,
%     breakable
% ]
% \textbf{System:} You are a careful annotator for a benchmark dataset.
% Output exactly one tag: \texttt{<answer>LABEL</answer>}.\\[0.5em]
% \textbf{User:} Task: classify the given excerpt into exactly one inconsistency type.\\[0.5em]
% Valid labels (choose exactly one), with a definition and one example each:\\
% Factual, Logical, Normative \& Policy, Numerical, Pragmatic, Referential,
% Specificity \& Scope, Temporal, Terminological, Theoretical,
% Unit \& Measurement.\\[0.5em]
% Return JSON with exactly this schema:\\
% \texttt{\{"label": "<one of the valid labels verbatim>"\}}\\[0.5em]
% Excerpt:\\
% \texttt{\{inconsistent\_text\}}
% \end{tcolorbox}

\section{Classification Prompt}
\label{app:prompt}
\begin{tcolorbox}[
    colback=gray!5,
    colframe=gray!50,
    title={\small\textbf{Classification Prompt (prompted Qwen / Gemma)}},
    fontupper=\small\ttfamily,
    breakable
]
\textbf{System:} You are a careful annotator for a benchmark dataset.
Output exactly one tag pair and nothing else: \texttt{<answer>LABEL</answer>}.\\[0.5em]
\textbf{User:} Task: classify the given excerpt into exactly one inconsistency type.\\[0.5em]
Valid labels (choose exactly one), with a definition and one example each:\\
Factual inconsistency, Logical inconsistency, Normative and policy inconsistency,
Numerical inconsistency, Pragmatic inconsistency, Referential inconsistency,
Specificity and scope inconsistency, Temporal inconsistency, Terminological inconsistency,
Theoretical inconsistency, Unit and measurement inconsistency.\\[0.5em]
Return EXACTLY (no other text): \texttt{<answer>LABEL</answer>}, where LABEL is copied verbatim from the list above.\\[0.5em]
Excerpt:\\
\texttt{\{inconsistent\_text\}}
\end{tcolorbox}

\end{document}